\documentclass[manuscript,screen]{acmart}

\usepackage{rotating}

\AtBeginDocument{%
  \providecommand\BibTeX{{%
    \normalfont B\kern-0.5em{\scshape i\kern-0.25em b}\kern-0.8em\TeX}}}

\setcopyright{acmcopyright}
\copyrightyear{2023}
\acmYear{2023}
\acmDOI{XXXXXXX.XXXXXXX}

\begin{document}

\title{Animal Behavior Analysis Methods Using Deep Learning: A Survey}

\author{Edoardo Fazzari}
\email{edoardo.fazzari@santannapisa.it}
\orcid{0000-0002-4570-4170}
\affiliation{%
  \institution{Sant'Anna School of Advanced Studies}
  \streetaddress{Piazza Martiri della Libertà}
  \city{Pisa}
  \country{Italy}
  \postcode{56127}
}

\author{Donato Romano}
\affiliation{%
  \institution{Sant'Anna School of Advanced Studies}
  \streetaddress{Piazza Martiri della Libertà}
  \city{Pisa}
  \country{Italy}
  \postcode{56127}
}
\email{donato.romano@santannapisa.it}

\author{Fabrizio Falchi}
\affiliation{%
  \institution{
  Institute of Information Science and Technologies, National Research Council of Italy}
  \streetaddress{Via G. Moruzzi}
  \city{Pisa}
  \country{Italy}
  \postcode{56124}
}
\affiliation{%
  \institution{Sant'Anna School of Advanced Studies}
  \streetaddress{Piazza Martiri della Libertà}
  \city{Pisa}
  \country{Italy}
  \postcode{56127}
}
\email{fabrizio.falchi@cnr.it}

\author{Cesare Stefanini}
\affiliation{%
  \institution{Sant'Anna School of Advanced Studies}
  \streetaddress{Piazza Martiri della Libertà}
  \city{Pisa}
  \country{Italy}
  \postcode{56127}
}
\email{cesare.stefanini@santannapisa.it}

\renewcommand{\shortauthors}{Fazzari, et al.}

\begin{abstract}
  Animal behavior serves as a reliable indicator of the adaptation of organisms to their environment and their overall well-being. Through rigorous observation of animal actions and interactions, researchers and observers can glean valuable insights into diverse facets of their lives, encompassing health, social dynamics, ecological relationships, and neuroethological dimensions. Although state-of-the-art deep learning models have demonstrated remarkable accuracy in classifying various forms of animal data, their adoption in animal behavior studies remains limited. This survey article endeavors to comprehensively explore deep learning architectures and strategies applied to the identification of animal behavior, spanning auditory, visual, and audiovisual methodologies. Furthermore, the manuscript scrutinizes extant animal behavior datasets, offering a detailed examination of the principal challenges confronting this research domain. The article culminates in a comprehensive discussion of key research directions within deep learning that hold potential for advancing the field of animal behavior studies.
\end{abstract}




\maketitle

\section{Introduction}
Animal behavior encompasses a spectrum of actions, reactions, and activity patterns demonstrated by animals in response to their environment, fellow organisms, and internal stimuli~\cite{Brown220855}. This expansive field covers a diverse range of behaviors, spanning from innate instincts and simple reflexes to intricate social interactions and learned conduct. The study of animal behavior involves the observation, description, and comprehension of how animals engage with one another and their surroundings. Presently, the landscape of animal behavior research is undergoing rapid evolution, propelled by the continual introduction of innovative experimental methodologies and the advancement of sophisticated behavior detection systems~\cite{wang2021behavioral}. This progression holds particular significance in advancing our understanding of neuroethological aspects, exemplified by the utilization of mice in exploring diseases like Alzheimer's~\cite{pedersen2006rosiglitazone}, and in refining animal welfare practices within agriculture~\cite{mishra2023advanced}. The impetus behind this surge in progress is the integration of cutting-edge technologies, with deep learning standing out as a transformative force that reshapes the approaches researchers employ to investigate and interpret animal behaviors\cite{Brown220855}.

Deep learning has emerged as a pivotal tool in the exploration of animal behavior. This advanced branch of artificial intelligence empowers computers to autonomously discern patterns and features from extensive datasets. As researchers amass increasingly intricate datasets through state-of-the-art monitoring technologies, including high-resolution cameras, GPS tracking devices, and sensors, the capability of deep learning algorithms to extract meaningful insights becomes indispensable~\cite{benaissa2023improved, koger2023quantifying}. This not only expedites the analysis process but also reveals nuanced aspects of animal behavior that were previously challenging to decipher.
Furthermore, deep learning plays a crucial role in the development of sophisticated behavior detection systems. These systems can automatically recognize and classify various behaviors, allowing researchers to redirect their focus from laborious manual data annotation to the interpretation of results~\cite{arablouei2023animal}. This acceleration in data processing and behavior recognition enhances the scalability and efficiency of animal behavior studies, ushering in a new era of discovery and understanding in this dynamic field.

\subsection{Survey structure}
The structure of our survey is systematically delineated as follows: In the initial section, we expound upon the underlying motivation propelling the use of deep learning for examining animal behaviors, explaining the inherent limitations therein. Concurrently, we articulate the research questions that our survey endeavors to address. Subsequently, a comprehensive exposition of the methodological framework employed for conducting the survey is presented, which includes a discerning analysis of the gathered data to elucidate discernible trends within the research domain. The subsequent segmentation of the study into two distinct components is pivotal to the overarching architecture of this article. These segments, namely pose estimation and non-pose estimation-based methods, constitute the primary constituents of our investigation, elucidating the extraction of salient information pertinent to animal behavioral analysis and its subsequent application in behavior identification. Following this delineation, we furnish a compendium of publicly available datasets. In conclusion, we revisit the initially posited research questions, providing responses in light of the findings expounded within the two principal segments of the article.

\subsection{Contributions}
This survey article makes a threefold contribution to the current understanding of the study of animal behavior through deep learning:

\begin{itemize}
    \item We provide a thorough examination of existing technologies and algorithms employed in the analysis of animal behavior. This entails a detailed exploration of methodologies and approaches currently prevalent in this domain, providing readers with a nuanced understanding of the technological landscape.
    \item We compile and present a comprehensive list of publicly available datasets relevant to the research field. This compilation serves as a valuable resource for researchers and practitioners, facilitating access to essential data for furthering investigations into animal behaviors through data-driven methodologies.
    \item We engage in a substantive discussion regarding potential directions for the evolution of the field. Emphasizing the integration of deep learning techniques, our discourse aims to enhance the quality of existing technologies, thereby advancing the understanding of animal behaviors. This forward-looking analysis provides insights into potential avenues for improvement and innovation.
\end{itemize}

To the best of our knowledge, this survey is the only examination of the topic to date. The comprehensive overview, dataset compilation, and forward-looking discussions collectively contribute to a nuanced understanding of current advancements and lay the groundwork for future developments in the application of deep learning to the study of animal behavior.

\section{Motivation and problem statement}
In this section, we expound upon the foundational motivations that underscore the study of animal behavior through deep learning, articulating the diverse advantages and objectives inherent to this specialized research domain. Concurrently, we meticulously scrutinize the principal limitations that characterize this field, recognizing the nuanced challenges that arise from variations in research setups. Our intention is to furnish a comprehensive guide for prospective researchers, endowing them with a thorough understanding of potential impediments prior to initiating investigations within this domain. Concurrently, we underscore the myriad opportunities inherent in the study of animal behavior, fostering an appreciation for the intricate dimensions of this research frontier. 

\begin{table}[ht]
  \centering
  \caption{Summary of the limitations and objectives in studying animal behaviors. DL stands for Deep Learning}
  \begin{tabular}{|p{5cm}|p{5cm}|}
    \hline
    \multicolumn{1}{|c|}{\textbf{Limitations}} & \multicolumn{1}{c|}{\textbf{Objectives}} \\
    \hline
    Sensor-induced stress & Biodiversity conservation \\
    Battery life & Biodiversity preservation \\
    Data noise & Ecological insight\\
    Storage constraints & Health impact\\
    Labeling economics & Welfare optimization\\
    Subjective annotation & Pest management\\
    Computational demands & Social dynamics\\
    Ethical considerations & Non-invasive (DL related) \\
    In-field tracking & Real-time applications (DL related) \\
    Environmental unpredictability &  \\
    \hline
  \end{tabular}
  \label{tab:limadv}
\end{table}

\subsection{Limitations in studying animal behaviors}
\label{ssec:limit}
The exploration of animal behavior is confronted by a multitude of challenges that intricately shape the effectiveness and practicality of its applications. These challenges permeate the methodologies employed for data acquisition, the intricacies of data analysis (both in terms of location and computational demands), and the nuanced process of data annotation.

An integral aspect of animal behavior studies is the utilization of sensors for data collection. However, the attachment of sensors to animals introduces a unique set of challenges. Notably, there is a risk of inducing stress responses and altering normal behaviors~\cite{zhang2020automated}. This necessitates a profound reflection on the authenticity of the collected data, urging researchers to question whether stress-induced behaviors accurately mirror natural patterns. Moreover, the task of differentiating genuine behavioral signals from background noise in sensor data adds complexity to interpretation, emphasizing the requirement for advanced algorithms capable of discerning meaningful patterns amidst the noise~\cite{kavlak2023disease}. Beyond stress responses and noise challenges, sensor equipment grapples with limitations in battery life~\cite{mekruksavanich2022resnet}, impacting the duration and scope of behavioral studies, especially in scenarios requiring continuous data collection over extended periods. Researchers are challenged to strike a balance between the need for comprehensive, continuous monitoring and the practical constraints imposed by limited battery capacities.

Transitioning into the domain of mobile devices introduces another layer of complexity to the challenges encountered in deep learning applications. The implementation of models on these devices confronts the persistent issue of storage limitations~\cite{cao2020real}. Balancing robust object detection with efficient data compression becomes a paramount concern, with techniques like Quantized-CNN~\cite{wu2016quantized} attempting to address this challenge. However, the ongoing quest is to achieve this balance without compromising precision, a crucial consideration for the reliability of behavioral analyses.

A pivotal challenge arises in the realm of labeling and annotation, where economic and practical constraints hinder the tagging of large datasets for each animal~\cite{bhattacharya2021pose}. This bottleneck impedes the scalability of deep learning models, heavily reliant on labeled datasets for effective training. The impracticality of manual labeling raises fundamental concerns about the breadth and accuracy of behavioral datasets, impacting the reliability of subsequent analyses. Subjectivity compounds these challenges, influencing the accuracy and consistency of behavioral annotations. Visual inspection, often subjective, is limited in providing objective insights into complex animal behaviors~\cite{bernardes2021ethoflow}. Manual annotation, while traditional, is labor-intensive and susceptible to inter-annotator disagreements~\cite{zhou2022cross, segalin2021mouse}. The inherent subjectivity introduces variability, raising questions about the replicability and reliability of experiments~\cite{hou2020identification, dell2014automated}.

Moreover, these challenges extend to innovative techniques, such as multi-view recordings, which hold promise for providing richer insights into animal behaviors~\cite{jiang2021multi}. However, challenges arise in correlating social behaviors from different perspectives due to the lack of correspondence across data sources. Effectively coordinating information from multiple viewpoints demands inventive approaches to ensure accuracy and reliability, representing a frontier where deep learning methodologies can contribute significantly.

A major challenge surfaces when comparing laboratory studies, where challenges often revolve around the subjectivity introduced by the controlled environment, with ethological studies conducted in the wild presenting a distinct set of challenges, notably in-field tracking~\cite{marshall2022leaving}. The diverse and unpredictable environments encountered in the wild introduce complexities not found in controlled laboratory settings. Bridging the gap between these two disciplines necessitates adaptable detection and tracking algorithms that seamlessly operate in both environments. Robust algorithms capable of handling varying animal sizes, changing appearances, clutter, occlusions, and unpredictable environments are vital for extracting meaningful insights~\cite{haalck2020towards, hou2020identification, lauer2022multi}. These challenges underscore the critical need for technological innovation that aligns with the demands of both controlled and wild settings.

\subsection{Objectives in studying animal behaviors}
Studying animal behavior offers manifold advantages, enriching our comprehension of the natural world and presenting practical applications across diverse domains, including neuroscience, pharmacology, medicine, agriculture, ecology, and robotics. Six key advantages of studying animal behavior are identified:

\begin{enumerate}
    \item \textbf{Biodiversity Conservation}: Understanding animal behavior is crucial for the conservation of biodiversity~\cite{hou2020identification, nilsson2020simple, wijeyakulasuriya2020machine, ditria2020automating, pillai2023deep}. Knowledge of behaviors such as migration patterns, feeding habits, and reproductive strategies is essential for designing effective conservation strategies and protecting endangered species.
    \item \textbf{Ecological Understanding:} Animal behavior provides insights into the ecological dynamics of ecosystems~\cite{akccay2020automated, gotanda2019animal}. Behavioral studies help researchers understand how animals interact with their environment, including their roles in nutrient cycling, seed dispersal, and predator-prey relationships~\cite{nasiri2023estimating, chen2020recognition, yamada2020evolution}.
    \item \textbf{Human Health and Medicine:} Studying animal behavior can have implications for human health and medicine~\cite{shaw2023human, hart2011behavioural, gnanasekar2022rodent, manduca2023learning}. For example, research on animal models helps in understanding certain diseases and developing potential treatments. Behavioral studies on animals also contribute to our understanding of the neurobiology and psychology that underlie human behavior~\cite{mathis2020deep, coria2022neurobiology, saleh2023pharmacology}.
    \item \textbf{Animal Welfare and Husbandry:}  Knowledge of animal behavior is essential for promoting the welfare of domesticated animals and optimizing their husbandry practices~\cite{jiang2021multi, bao2022artificial}. Understanding how animals express natural behaviors can inform the design of environments that support their physical and psychological well-being~\cite{tassinari2021computer, manoharan2020embedded, jiang2020automatic}.
    \item \textbf{Pest Control and Agriculture:} Although right now it is very limited to pest identification, understanding the behavior of pest species can aid in the development of effective pest control strategies in agriculture~\cite{coulibaly2022explainable, junior2020automatic, mendoza2023application}. This knowledge helps farmers manage crop damage and reduce the need for harmful pesticides~\cite{teixeira2023systematic, mankin2021automated}.
    \item \textbf{Understanding Social Dynamics:} Observing social behaviors in animals can provide insights into the principles governing social structures and interactions~\cite{xiao2023multi, papaspyros2023predicting}. This knowledge can have applications in fields such as sociology and psychology, contributing to our understanding of social dynamics in general~\cite{alameer2022automated, perez2023cnn, landgraf2021animal}.
\end{enumerate}

In this context, deep learning plays a pivotal role and emerges as a major technology in advancing the field, opening new opportunities. Addressing the limitations discussed in the previous section, we will now elaborate on the advantages that deep learning and computational technologies bring to the study of animal behavior.

We previously emphasized the integral aspect of employing sensors for data collection in animal behavior studies, which may induce stress and high noise levels. The use of multiple and diverse sensors for data acquisition, coupled with advanced architectures incorporating fusion layers, has been shown to mitigate noise and enhance the precision of analysis~\cite{mahmud2021systematic}. However, attaching sensors directly to animals may introduce bias, prompting researchers in livestock health assessment and neuroscience to adopt computer vision. The ability of computer vision to provide real-time, non-invasive, and accurate animal-level information through the use of cameras has gained popularity~\cite{oliveira2021review}. Nevertheless, this approach is limited to setups within the camera frame, except for innovations like Haalck \textit{et al.}'s~\cite{haalck2020towards} moving camera that tracks animals, creating a dynamic map of their environment. In larger scenarios, such as meadows where cows graze, sensors remain preferable. Nevertheless, collecting and analyzing sensor data from mobile devices on animals proves challenging and time-consuming. To address this, Dang \textit{et al.}~\cite{dang2022lorawan} introduced Long Range Area Network (LoRaWAN), where sensors attached to cows connect to gateways transmitting information to the cloud. This not only overcomes the limitations of computational power associated with mobile devices but also ensures continuous, real-time data analysis.

The efficacy of deep learning is contingent on annotated data, especially for supervised approaches. While manual annotation remains unavoidable, in tasks such as pose estimation and classification, labeling can be iterative. This involves annotating a small portion of the dataset, training a network, predicting on new images, correcting labels, and repeating this process multiple times. This iterative approach accelerates the labeling process, as demonstrated by Pereira \textit{et al.}~\cite{pereira2019fast}. Another approach is to generate artificial labels~\cite{li2023scarcenet}.

\autoref{tab:limadv} succinctly encapsulates a consolidated overview of both primary limitations and advantages, providing a discerning reference for researchers navigating the sophisticated landscape of animal behavior studies.

\subsection{Research questions}
\label{ssec:rq}
The survey aims to address the following research questions:

\textbf{RQ1} \textit{Which animal species are less considered and why?}

\noindent This research question seeks to investigate overlooked or less studied animal species in the context of behavior analysis. To address this question, the survey will explore existing literature and research to identify trends and biases in the selection of animal subjects. The objective is to understand why certain species receive less attention and to gain insights into potential gaps in knowledge, aiding the development of more comprehensive and inclusive research strategies.

\textbf{RQ2} \textit{What deep learning methods have been used in the literature for animal behavior analysis?}

\noindent This research question focuses on summarizing and categorizing the existing deep learning methods employed in the literature for animal behavior analysis. The survey will review a wide range of studies to identify and classify the various deep learning techniques applied to analyze animal behavior. The objective is to analyze existing methodologies to identify trends, strengths, and limitations of current approaches in the field.

\textbf{RQ3} \textit{What are the differences between human and animal behavior analysis?}

\noindent This research question aims to highlight the distinctions between the analysis of human behavior and that of other animal species. The survey will compare methodologies, ethical considerations, and challenges specific to studying human and animal behavior.

\textbf{RQ4} \textit{What are the deep learning strategies that are suitable and could enhance this task, but are not yet exploited?}

\noindent This research question looks forward, aiming to identify untapped potential in the application of deep learning to animal behavior analysis. The survey will involve a comprehensive review of the current literature to identify gaps or areas where deep learning strategies have not been extensively explored. This involves proposing novel applications of existing techniques or suggesting modifications to adapt deep learning methods for more effective analysis of animal behavior.


\section{Method for literature survey}
In this section, we clarify the methodology applied in our survey. Our approach involved a thorough systematic review to carefully select the pertinent studies considered in this article. Following this, we accurately analyzed the gathered information, employing statistical methods to derive meaningful insights.

\subsection{Search and selection strategies}
This section delineates the methodologies employed for data collection and synthesis. Initially, data acquisition was conducted through systematic searches on academic repositories, including Google Scholar, IEEE Xplore, and the Springer Database. The formulated search queries were as follows:

\begin{center}
    \textit{animal behavior} \textbf{AND} \textit{deep learning}

    (\textit{insect} \textbf{OR}  \textit{wild}) \textbf{AND}  \textit{behavior} \textbf{AND}  \textit{deep learning}
\end{center}

The decision to employ distinct queries for insects and wild animals was necessitated by the observed paucity of literature in these categories relative to studies involving farm animals and neuroethology, commonly focused on mice. Thus, the formulation of specific queries was imperative to encompass a broader spectrum of animal species.

Subsequently, the acquired data underwent systematic tabulation based on features explicated in \autoref{tab:paperfeatures}. These features were derived from discerned patterns identified during a comprehensive analysis of the extant literature. Synthesizing the outcomes of this analysis, Sections \ref{sec:pose} and \ref{sec:nopose} encapsulate the aggregated findings, summarizing the respective papers that expound upon solution methodologies grounded in pose estimation and those that do not.

Finally, a judicious filtering operation was executed to extract only those articles germane to the objectives of this survey, resulting in 151 articles. Each article within this subset underwent thorough examination, and pertinent references therein were scrutinized and subsequently incorporated into our survey to enrich its content.

\begin{table}[ht]
    \caption{Features Extracted from Papers}
    \begin{tabular}{p{0.9\linewidth}}
        \hline
        \textbf{Reference:} Assigned identifier for each retrieved article. \\
        \textbf{Year:} Publication year of the article. \\
        \textbf{Country:} Geographical location where the authors are based, as required in Section \ref{ssec:mappinganalysis}. \\
        \textbf{Species:} The specific species under investigation in the article. \\
        \textbf{Pose Estimation:} Indicates whether the methodology incorporates pose estimation (\textit{True} or \textit{False}). \\
        \textbf{Behavior Analysis:} Indicates whether the analysis considers an association between extracted features and observed behaviors (\textit{True} or \textit{False}). \\
        \textbf{Feature Methodology:} The approach employed to extract salient features in the study. \\
        \textbf{Behavior Methodology:} The methodological framework used to correlate features with observed behaviors. \\
        \textbf{Authors' research field:} Indicates the research fields the authors mainly work on. \\
        \hline
    \end{tabular}  
    \label{tab:paperfeatures}
\end{table}

\subsection{Comprehensive science mapping analysis}
\label{ssec:mappinganalysis}

\subsubsection{Annual scientific production} In the process of retrieving articles, our attention was exclusively directed towards research publications spanning the temporal spectrum from 2020 to 2023. \autoref{fig:analysis}\textbf{a} elucidates this distribution through a histogram, illustrating the quantitative representation of papers across each respective year.

\subsubsection{Scientific production based on animal considered} \autoref{fig:analysis}\textbf{b} illustrates the distribution of percentages pertaining to the various animal species under consideration in the selected articles. Evidently, a predominant emphasis is placed on research concerning livestock, notably focusing on cows and pigs, as well as studies involving mice, related mostly to neuroscience. 


\subsubsection{Research field of authors}
\begin{figure}[ht]
    \centering
    \includegraphics[width=0.99\linewidth]{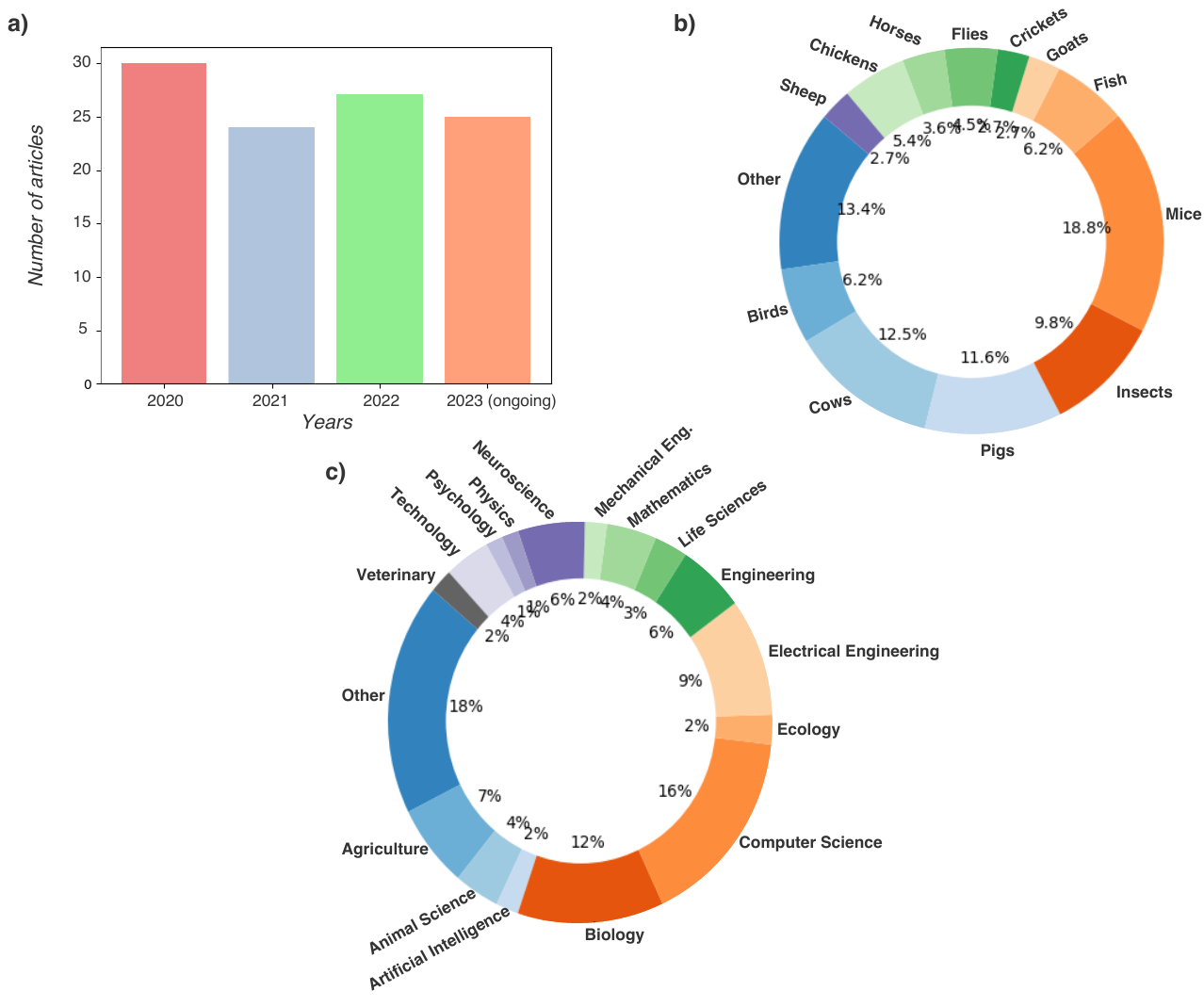}
    \caption{\textbf{a)} illustrates a histogram depicting the distribution of research articles per year, focusing exclusively on papers obtained and cataloged during the initial scavenging phase. \textbf{b)} presents a pie chart detailing the variety of animals utilized in behavioral studies leveraging deep learning techniques. \textbf{c)} displays another pie chart showcasing the diverse research fields of the authors.}
    \label{fig:analysis}
\end{figure}

Given that animal behavior analysis is an interdisciplinary and multidisciplinary field, it becomes imperative to comprehend the research background of individuals engaged in this domain. Despite the predominant focus on articles related to deep learning technologies, it is noteworthy that a considerable number of non-artificial intelligence practitioners are actively entering this field, as illustrated in \autoref{fig:analysis}\textbf{c}.  Interestingly, when combining "computer science" (encompassing computer engineering) and "artificial intelligence," they constitute only 18\% of the scholarly contributions. In contrast, bio-related fields, including biology, animal science, agriculture, veterinary, and ecology, collectively contribute 30\% to the research landscape.
Noteworthy is the active participation of various engineering fields, even those with a mechanical-electrical background, in the exploration of animal behavior. Additionally, a compelling correlation is observed in the fields of neuroscience and psychology, where the majority of articles are dedicated to the study of mice. 

\section{Pose estimation-based methods}
\label{sec:pose}
Pose estimation, the process of identifying and locating the position and orientation of objects, is a fundamental technique widely used in the examination of animal behaviors alongside object detection, as discussed in \autoref{ssec:objectdetection}. Originating from Human Pose Estimation (HPE), the evolution into Animal Pose Estimation (APE) was spearheaded by Mathis et al. through DeepLabCut~\cite{mathis2018deeplabcut} and Pereira et al. via LEAP~\cite{pereira2019fast}, subsequently evolving into SLEAP~\cite{pereira2022sleap}. This section delves into an in-depth analysis of these two methodologies juxtaposed with emerging trends within the field of research. Given the primary focus of our survey on animal behavior analysis, subsequent to the introduction of these predominant approaches, we elucidate the utilization of pose estimation outputs for behavior analysis and classification. For a more comprehensive understanding of animal pose estimation, we recommend perusing the survey conducted by Jiang et al.~\cite{jiang2022animal}, published in 2022.

\begin{figure}[ht]
    \centering
    \includegraphics[width=0.9\linewidth]{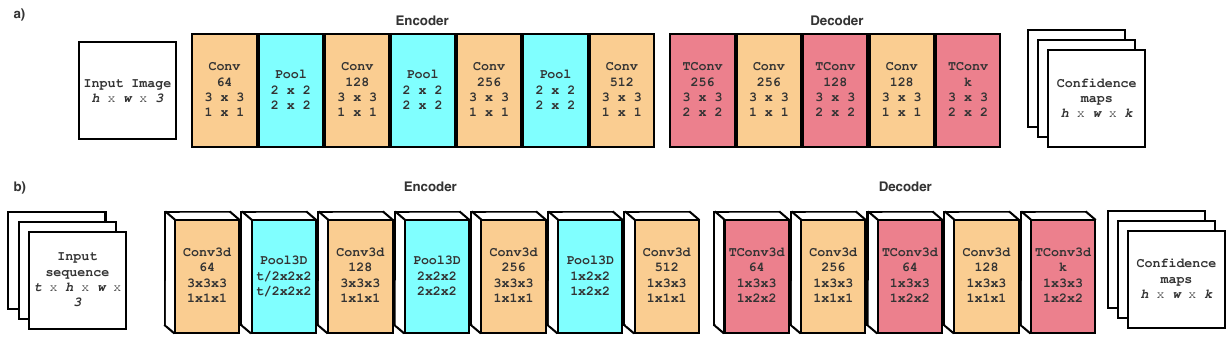}
    \caption{\textbf{a)} shows the architecture of LEAP~\cite{pereira2019fast}; \textbf{b)} the one exploited in T-LEAP~\cite{russello2022t}}
    \label{fig:leap}
\end{figure}

LEAP~\cite{pereira2019fast} is a single-animal pose estimation model employing convolutional layers culminating in confidence maps that delineate the probability distribution for each distinct body keypoint. This architectural design, depicted in \autoref{fig:leap}, is characterized by its simplicity, featuring three sets of convolutional layers. The initial two sets are terminated by \textit{max pooling} to alleviate computational complexity. Subsequently, transposed convolution is applied to restore the original dimensions of the images, yet with a depth corresponding to the number of keypoints, thereby generating a confidence map for each. Despite its simplicity, the LEAP model encounters challenges in non-laboratory settings due to issues such as occlusion, prompting the introduction of T-LEAP~\cite{russello2022t}. T-LEAP preserves the architecture of LEAP but diverges in its use of 3D convolution instead of 2D convolution. The input to T-LEAP comprises four consecutive frames extracted from videos, enhancing the model's robustness. Notably, T-LEAP maintains a focus on single-animal pose estimation, as elucidated in \autoref{fig:leap}. 
Subsequently, the author of LEAP introduced a refined version known as Social LEAP (SLEAP)~\cite{pereira2022sleap}, designed to proficiently address the challenges associated with multi-animal pose estimation through the integration of both bottom-up~\cite{Papandreou_2018_ECCV} and top-down strategies~\cite{nguyen2022survey}. In the top-down strategy, SLEAP first identifies individuals and subsequently detects their respective body parts. Unlike LEAP, SLEAP seamlessly incorporates this approach without the need for an additional object detection architecture. On the other hand, the bottom-up strategy in SLEAP involves detecting individual body parts and subsequently grouping them into individuals based on their connectivity. A key advantage of this dual-strategy framework is its efficiency, requiring only a single pass through the neural network. The output of this strategy produces multi-part confidence maps and part affinity fields (PAFs)~\cite{cao2017realtime}, constituting vector fields that intricately represent spatial relationships between pairs of body parts. Additionally, SLEAP undergoes a structural enhancement by transitioning from LEAP's backbone to a more intricate U-Net architecture~\cite{ronneberger2015u}, thereby significantly improving accuracy in the realm of multi-animal pose estimation scenarios.

Similarly, DeepLabCut (DLC)~\cite{mathis2018deeplabcut} has evolved significantly over time. Initially designed as a single-animal pose estimation method, it utilizes a pretrained ResNet-50~\cite{he2016identity} backbone with subsequent deconvolutional layers to generate confidence maps for keypoints. This approach, taking advantage of Imagenet pretrained weights, allowed DLC to effectively estimate skeletons with minimal data. The model's capabilities were later expanded to include 3D pose estimation through the use of multiple cameras~\cite{nath2019using}. Each camera view was trained independently, and sophisticated camera calibration techniques were employed to derive 3D locations. A subsequent milestone in DLC's development involved addressing the challenges of multi-animal pose estimation~\cite{lauer2022multi}. This evolution introduced DLCRNet, a structural modification that replaced the ResNet backbone. DLCRNet employs a bottom-up multi-animal pose estimation approach, featuring a multi-fusion architecture and a multi-stage decoder. The decoder utilizes multiple stages of score maps and PAFs~\cite{cao2017realtime} to predict keypoints for each animal. Further innovation is exemplified by SuperAnimal~\cite{ye2022superanimal}, which introduced transformer layers into the model architecture.

While DLC~\cite{mathis2018deeplabcut} and SLEAP~\cite{pereira2022sleap} currently stand as the predominant pose estimation methodologies in behavior analysis for animal behavior classification, it is imperative to acknowledge recent advancements in animal pose estimation architectures. Several notable methodologies have been introduced:

\begin{itemize}
    \item \textbf{OptiFlex}~\cite{liu2020optiflex} is a video-based animal pose estimation method that, given a skip ratio \(s\) and a frame range \(f\), assembles a sequence of \(2f + 1\) images with indices ranging from \(t - s \times f\) to \(t + s \times f\). This sequence is input to a model based on residual blocks with intermediate supervision, generating predictions for each image and producing a sequence of heatmap tensors. These tensors are then fed into an OpticalFlow model, ultimately yielding the final heatmap prediction for index \(t\). OptiFlex has demonstrated superior accuracy compared to DeepLabCut~\cite{mathis2018deeplabcut}, LEAP~\cite{pereira2019fast}, and DeepPoseKit~\cite{graving2019deepposekit}.

    \item \textbf{SemiMultiPose}~\cite{blau2022semimultipose} introduces a semi-supervised multi-animal pose estimation approach, building upon DeepGraphPose~\cite{wu2020deep} and DirectPose \cite{tian2019directpose}. Taking both labeled and unlabeled frames as input, the method processes them using a ResNet~\cite{he2016identity} backbone, generating a compact representation fed into three branches: one for detecting keypoint heatmaps (B1), one for bounding box heatmaps (B2), and a third for keypoint detection (B3). SemiMultiPose aims to generate pseudo keypoint coordinates from B2 and B3 for the self-supervised branch, contributing to B1. The network has shown improved accuracy compared to SLEAP~\cite{pereira2022sleap}. However, the authors note that in cases of abundant labeled data, their method may not significantly outperform others, and for single-animal pose estimation with unlabeled frames from a sequential video, DeepGraphPose~\cite{wu2020deep} might outperform SemiMultiPose~\cite{blau2022semimultipose}, benefiting from the consideration of spatial and temporal information.

    \item \textbf{Lightning Pose} \cite{biderman2023lightning} exploits spatiotemporal statistics of unlabeled videos in two ways. Firstly, it introduces unsupervised training objectives penalizing the network for predictions violating the smoothness of physical motion, multiple-view geometry, or departing from a low-dimensional subspace of plausible body configurations. Secondly, it proposes a novel network architecture predicting poses for a given frame using temporal context from surrounding unlabeled frames. The resulting pose estimation networks exhibit superior performance with fewer labels, generalize effectively to unseen videos, and provide smoother and more reliable pose trajectories for downstream analysis (e.g., neural decoding analyses) compared to previously mentioned approaches.
    
    \item \textbf{Bhattacharya \textit{et al.}} \cite{bhattacharya2021pose} introduced a novel model for recognizing the pose of multiple animals from unlabeled data. The approach involves the removal of background information from each image and the application of an edge detection algorithm to the body of the animal. Subsequently, the motion of the edge pixels is tracked, and agglomerative clustering is performed to segment body parts. In a departure from previous methods, the end result is not specific keypoints but rather the segmentation of body parts. To achieve this, the authors utilized contrastive learning to discourage the grouping of distant body parts together.
\end{itemize}

After obtaining the skeletal representation of each animal in every frame, whether from videos or images, the subsequent step involves processing the data to discern specific behaviors. The trajectories derived from pose estimation can be effectively analyzed through statistical methods. Weber \textit{et al.}~\cite{weber2022deep} utilized DeepLabCut predictions~\cite{mathis2018deeplabcut} and ANOVA~\cite{kaufmannANOVA} to conduct behavioral profiling of rodents, with a focus on studying stroke recovery. In a similar vein, Lee \textit{et al.}~\cite{lee2021predicting}, employing DLC~\cite{mathis2018deeplabcut}, investigated the behavior of non-tethered fruit flies. Their study involved predicting locomotion patterns and identifying the centroid of the animals' legs.

Machine learning for analyzing pose estimation trajectories becomes crucial when classifying postures and relating them to specific behaviors. One of the simplest approaches is to use a Nearest-Neighbor classifier. Saleh \textit{et al.}~\cite{saleh2023pharmacology} tested this method to classify mouse behaviors such as crossing and rearing in an open-field experiment, achieving 97\% accuracy. Other machine learning approaches were employed by Fang \textit{et al.}~\cite{fang2021pose} and Nilsson \textit{et al.}~\cite{nilsson2020simple}. The former used a naive Bayesian classifier to identify eating, preening, resting, walking, standing, and running behaviors for poultry analysis, providing a disease warning system. The latter introduced the SimBa toolkit, importing DeepLabCut~\cite{mathis2018deeplabcut} or DeepPoseKit~\cite{graving2019deepposekit} projects to create classifiers using RandomForest~\cite{breiman2001random} and extracting features like velocities and total movements. McKenzie-Smith \textit{et al.}~\cite{mckenzie2023capturing} used trajectories obtained with SLEAP~\cite{pereira2022sleap} to identify stereotyped behaviors such as grooming, proboscis extension, and locomotion in \textit{Drosophila melanogaster}, using resulting ethograms provided by MotionMapper~\cite{berman2014mapping} to explore how flies' behavior varies across time of day and days, finding distinct circadian patterns in all stereotyped behaviors.

Other authors opted for recurrent and convolutional neural networks, with simple approaches such as using Long Short-Term Memory (LSTM)~\cite{hochreiter1997long} and 1D convolutional neural networks to process trajectories for drawing behavioral conclusions. Examples include detecting lameness in horses ~\cite{alagele2022animal} and determining chemical interactions experienced by crickets ~\cite{fazzari2}. More complex approaches include Wittek \textit{et al.}~\cite{wittek2023supervised}'s use of InceptionTime~\cite{ismail2020inceptiontime}, an ensemble of deep convolutional neural network models, to classify seven distinct behaviors in birds. Some authors simplified the classification process by introducing a non-linear clustering phase to improve the feature space, followed by classification using Multilayer Perceptrons (MLP), demonstrating advantages in classification~\cite{ye2022superanimal, schneider2023learnable}.

A recent emerging trend involves the utilization of unsupervised learning techniques in the analysis of animal behavior. Luxer \textit{et al.}~\cite{luxem2022identifying} have innovatively proposed a methodology for processing trajectories derived from DeepLabCut~\cite{mathis2018deeplabcut} by employing a Variational Auto-Encoder (VAE)~\cite{kingma2013auto}. Subsequently, they apply a Hidden Markov Model (HMM)~\cite{rabiner1986introduction} to the new representation of trajectories to discern underlying motifs. Following a comprehensive analysis of motif usage, the authors iteratively employ HMM, limiting the number of motifs to those surpassing a 1\% usage threshold in the previous analysis. The refined motifs were attributed to specific behavior exhibited by the mice, such as exploration, rearing, grooming, pausing, or walking. Notably, this methodological approach outperforms conventional techniques, such as Auto-Regressive HMM (AR-HMM) or MotionMapper~\cite{berman2014mapping}, when applied directly to the motion sequences.

Motion trajectories extend their utility beyond predicting the behavior of individual animals; in multi-animal scenarios, they can also be applied to unravel the intricate web of social interactions among them. Segalin \textit{et al.}~\cite{segalin2021mouse} introduced the Mouse Action Recognition System (MARS), a sophisticated automated pipeline tailored for pose estimation and behavior quantification in pairs of freely interacting mice. MARS adeptly discerns three specific social behaviors: close investigation, mounting, and attack. On a different note, Zhou \textit{et al.}~\cite{zhou2022cross} proposed the Cross-Skeleton Interaction Graph Aggregation Network (CS-IGANet), a groundbreaking framework designed to capture the diverse dynamics of freely interacting mice. CS-IGANet successfully identifies a spectrum of behaviors, including approaching, attacking, chasing, copulation, walking away from another mouse, sniffing, and many others.

Trajectories not only serve as a means to identify specific behaviors but are also instrumental in anomaly detection. For instance, Fujimori \textit{et al.}~\cite{fujimori2020animal} employed OneClassSVM~\cite{boser1992training} and IsolationForest~\cite{liu2008isolation} to detect outlier behaviors in domestic cats. Similarly, Gnanasekar \textit{et al.}~\cite{gnanasekar2022rodent} utilized pose estimation data to predict abnormal behavior in mice undergoing opioid withdrawal, employing pretrained convolutional neural networks for the classification of shaking behaviors.

\section{Non pose estimation-based methods}
\label{sec:nopose}
In this section, we expound upon methodologies employed in the investigation of animal behaviors without recourse to pose estimation techniques. To enhance clarity and systematic presentation, we have delineated subsections corresponding to each methodology.

\subsection{Sensor based approaches}
Sensor-generated data, typically originating from accelerometers or gyroscopes, has been extensively explored in the literature, as comprehensively in~\cite{kleanthous2022survey, neethirajan2020role} surveys. These surveys delve into the application of classical machine learning methods in modern animal farming and the study of animal behavior. More recently, a shift towards leveraging deep learning approaches has been observed. Arablouei \textit{et al.}~\cite{arablouei2023animal} utilized a wearable collar tag equipped with an accelerometer to collect data from grazing beef cattle. They applied a Multi-Layer Perceptron to classify behaviors such as grazing, walking, ruminating, resting, and drinking. Similarly, Eerdekens \textit{et al.}~\cite{eerdekens2020resampling} employed tri-axial accelerometers on horses, strategically positioned at the two front legs' lateral side. They proposed a Convolutional Neural Network to detect behaviors like standing, walking, trotting, cantering, rolling, pawing, and flank-watching based on the acquired data. Mekruksavanich \textit{et al.}~\cite{mekruksavanich2022resnet}, instead, segmented accelerometer data into 2-second windows and exploited a pre-trained ResNet model~\cite{he2016identity} to perform sheep activity recognition. Dang \textit{et al.}~\cite{dang2022lorawan} introduced the integration of multiple sensors, collecting environmental data (e.g., temperature, humidity) alongside cow behavior information obtained from accelerometers and gyroscopes. They preprocessed this information using a 1D-convolutional neural network and LSTM networks for classifying walking, feeding, lying, and standing. In a recent study, Pan \textit{et al.}~\cite{pan2023cnn} introduced four novel Convolutional Neural Network architectures tailored for Animal Action Recognition (AAR). These architectures, namely one-channel temporal (OCT), one-channel spatial (OCS), OCT and spatial (OCTS), and two-channel temporal and spatial (TCTS) networks, leverage data from 3D accelerometers and 3D gyroscopes. The core objective of their research was to mscrupulously identify behaviors such as movement, drinking, eating, nursing, sleeping, and lying in lactating sows.

In addition to accelerometer and gyroscope data, GNSS (Global Navigation Satellite System) data emerges as a valuable tool for understanding animal behavior. Arablouei \textit{et al.}~\cite{arablouei2023multimodal} explored this avenue by employing GNSS to extract pertinent information about cattle behavior, including metrics like distance from water points, median speed, and median estimated horizontal position error. Integrating this GNSS data with accelerometry information, the researchers pursued two distinct approaches. The first involved concatenating features from both sensor datasets into a comprehensive feature vector, subsequently fed into a MLP classifier. Alternatively, the second approach centered on fusing the posterior probabilities predicted by two separate MLP classifiers. These methodologies enabled the accurate detection of behaviors such as grazing, walking, resting, and drinking.

\subsection{Bioacoustics}
While bioacoustics offers a captivating glimpse into animal behavior~\cite{stowell2022computational}, given the integral role of sound in animal activities such as communication, mating, navigation, and territorial defense~\cite{chalmers2021modelling}, the current landscape of published articles predominantly emphasizes animal identification~\cite{xu2020multi, bravo2021bioacoustic, varma2021acoustic} and sound event detection~\cite{nolasco2023learning, moummad2023regularized}. Notably, the existing literature reveals a scarcity of research endeavors combining acoustics and deep learning for the identification of animal behaviors. Wang \textit{et al.}~\cite{wang2021identification} stand out as pioneers in this domain, as they endeavored to classify sheep behaviors, including chewing, biting, chewing-biting, and ruminating sounds. This was accomplished using a recording device positioned proximal to the animal's face, with a placebo class designated as noise. 
The acquired wavelet data were leveraged for classification tasks through both a feed-forward neural network and a recurrent neural network. Additionally, the information was further processed by transforming it into a log-scaled Mel-spectrogram, serving as input for a convolutional neural network. The findings underscore that while the recurrent neural network exhibited superior performance, the convolutional neural network outperformed the feed-forward approach, attributing its success to the enhanced signal representation offered by the Mel-spectrogram.

\subsection{Object Detection}
\label{ssec:objectdetection}

In conjunction with pose estimation techniques, object detection stands out as a widely employed deep learning methodology for analyzing animal behavior. Its prevalence may be attributed to its established utility in animal recognition and detection~\cite{chen2021behaviour, teixeira2023systematic, banerjee2023deep}, prompting researchers to redirect their focus toward studying animal welfare and activity.

Among the leading architectures for animal behavior identification, Faster R-CNN~\cite{ren2015faster} and particularly YOLO~\cite{redmon2016you} are frequently employed. Nonetheless, alternative architectures have been proposed. For instance, Samsudin \textit{et al.}~\cite{samsudin2022zebrafish} utilized SSD MobileNetv2~\cite{sandler2018mobilenetv2} to detect abnormal and normal zebrafish larvae behaviors for examining the effects of neurotoxins. McIntosh \textit{et al.}~\cite{mcintosh2020movement} introduced TempNet, incorporating an encoder bridge and residual blocks with a two-staged spatial-temporal encoder, to detect startle event in fish.

Object detection serves a dual role, encompassing instantaneous behavior detection through image or video frame analysis, as well as the quantification and tracking of specific behaviors. The accurate analysis of single frames, counting, and frame-by-frame examination enable researchers to quantify both the duration and frequency of distinct actions. For instance, the application of YOLO in the study by Alameer \textit{et al.}~\cite{alameer2022automated} facilitated the quantification of contact frequency among pigs, allowing the identification of peculiar behaviors such as rear snorting and tail-biting. In the context of cows and pigs, a crucial aspect involves quantifying movement and aggressive behavior~\cite{alameer2022automated, odo2023video}. Furthermore, efforts to discern rank relationships based on fighting behavior in animals like cows are of crucial importance~\cite{uchino2021individual}. Importantly, for tasks demanding prolonged animal identification, tracking is conventionally executed using DeepSort~\cite{wojke2017simple, evangelista2022yolov7}.

Efficient instant detection can be accomplished by conducting a single analysis on the animal and directly classifying its behavior through a single image. In this context, deep learning object detection models prove instrumental in directly identifying behaviors such as positional activities (e.g., mating, standing, feeding, spreading, fighting, drinking) for the comprehensive analysis of animal health and stress behaviors~\cite{RIEKERT2020105391, manoharan2020embedded, wang2020real}. These models also find application in disease identification, such as the detection of wryneck~\cite{elbarrany2023use}, and in studying behavioral adaptations to new environments~\cite{li2019detection}. Furthermore, object detection models can be extended to operate with thermal and infrared images. For example, Xudong \textit{et al.}~\cite{xudong2020automatic} utilized thermal images for the automatic recognition of dairy cow mastitis, introducing the EFMYOLOv3 model. Similarly, Lei \textit{et al.} employed infrared images to discern feeding, resting, moving, and socializing behaviors in slow animals~\cite{lei2022postural}.

Beyond these applications, notable approaches utilizing object detection include Fuentes \textit{et al.}~\cite{fuentes2020deep}, who integrated YOLO and Optical Flow to detect actions in cows. Additionally, some researchers employ object detection solely for localizing the animal within the image or video. They subsequently crop that region and use it in other models, leveraging 2D pretrained networks or introducing 3D convolutional neural networks for video analysis~\cite{feighelstein2023deep, thanh2022deep}.

\subsection{Others}
Several research endeavors have employed unique deep learning methodologies, distinct from those discussed in the preceding section. Due to the relative scarcity of deep learning strategies for evaluating animal behavior in the existing literature using the aforementioned approaches, we endeavored to compile a comprehensive assortment of ideas. To achieve this, we have identified and categorized five distinct approaches:

\begin{itemize}
    \item \textbf{Convolutional Classification on Raw Data.} Alameer \textit{et al.}~\cite{alameer2020automatic} employed a GoogLeNet-like architecture\cite{szegedy2015going} to discern between feeding and non-nutritive visits to a manger in pig recordings. In a similar vein, Ayadi \textit{et al.}~\cite{ayadi2020dairy} utilized VGG19~\cite{simonyan2014very} to determine whether cows were ruminating or not. This network architecture was also applied to identify various behaviors in mice, such as grooming, licking the abdomen, squatting, resting, circling, wandering, climbing, and searching~\cite{wang2021behavioral}. Similarly leveraging pre-trained networks, Andresen \textit{et al.}~\cite{andresen2020towards} developed a fully automated system for surveilling post-surgical and post-anesthetic effects in mice facial expressions, employing InceptionV3~\cite{szegedy2016rethinking} for pain identification. Notably, Bohnslav \textit{et al.}~\cite{bohnslav2021deepethogram} introduced DeepEthogram, a software tested for predicting mice and flies behaviors. The approach involves using a sequence of 11 frames, where the last frame is the target for prediction. Optical flow frames are generated using MotionNet~\cite{zhu2019hidden}. These frames, along with the target frame, are fed into a feature extractor (ResNet architectures~\cite{he2016identity, hara2018can}) to extract both flow and spatial features. Subsequently, the features are concatenated, and Temporal Guassian Mixture (TGM) model~\cite{piergiovanni2019temporal} is applied for classification. Han \textit{et al.}~\cite{han2020fish} employed a simpler approach, superimposing the frame to be predicted with computationally generated optical flow from the subsequent frame. The resulting image is then classified using a convolutional neural network to categorize behaviors in fish shoals, including normal state, group stimulated, individual disturbed, feeding, anoxic, and starvation state.

    \item \textbf{Segmentation.} Xiao \textit{et al.}\cite{xiao2023multi} employed Mask R-CNN~\cite{he2017mask} to segment birds within a 3D space, facilitating the analysis of their interactions based on distinct social actions: approach, stay, leave, and sing to. In contrast, other researchers have devised innovative pipelines to investigate animal behavior. EthoFlow\cite{bernardes2021ethoflow} is a software grounded in segmentation, enabling the tracking and behavioral analysis of organisms (validated on bee datasets). On a different note, SIPEC~\cite{marks2022deep} constitutes a pipeline leveraging an Xception network~\cite{chollet2017xception} to extract features from frames. These features are subsequently processed over time using a Temporal Convolutional Network (TCN)~\cite{lea2016temporal} to classify the animal's behavior in each frame. While SIPEC abstains from segmentation in the case of single animal classification, it seamlessly incorporates segmentation for multi-animal behavior classification.
    
    \item \textbf{Self-Supervised Learning.} Jia \textit{et al.}~\cite{jia2022selfee} proposed an innovative self-supervised learning approach known as Selfee, designed for extracting comprehensive and discriminative features directly from raw video recordings of animal behaviors. Selfee utilizes a pair of Siamese convolutional neural networks\cite{koch2015siamese}, trained explicitly to generate discriminative representations for live frames.

    \item \textbf{Explainability.} To the best of our knowledge, Choi \textit{et al.}~\cite{choi2022beyond} stands as the sole contributor employing Explainable Artificial Intelligence (XAI). In their study, they harnessed Grad-CAM~\cite{selvaraju2016grad} to delve into the decision-making process of a neural network designed to distinguish between unstable and stable ant swarms. The investigation aimed to ascertain the network's capacity to comprehend intricate behaviors such as dueling and dominance biting, shedding light on the explainability of its predictions.

    \item \textbf{Behavior identification in clips.} Li \textit{et al.}~\cite{li2020spatiotemporal} undertook the task of categorizing significant pig behaviors, such as feeding, lying, motoring, scratching, and mounting. Their approach involved the development of Pig's Multi-Behavior recognition (PMB-SCN), a sophisticated architecture built upon the SlowFast framework~\cite{feichtenhofer2019slowfast} and leveraging spatio-temporal convolution. PMB-SCN comprised two distinct SlowFast pathways with varying temporal speeds. The \textit{slow pathway} utilized a larger temporal stride when processing input frames (e.g., 8, considering a clip with a length of 64 frames), while the \textit{fast pathway} employed a smaller temporal stride (e.g., 2). The features extracted by these pathways were interconnected through lateral connections~\cite{lin2017feature}, enhancing the model's ability to capture complex spatio-temporal patterns. The final phase of the methodology involved classification, where the fused features were utilized to discern and categorize various pig behaviors effectively.
\end{itemize}

\section{Publicly available datasets}
This section meticulously enumerates publicly accessible datasets featured or referenced in the articles identified through our systematic search. \autoref{tab:datasets} presents details on each dataset, including the article of introduction, authorship, targeted species, data type (e.g., images, videos, audio signals, sensor data), the specific tasks for which they were utilized and the content of the datasets.
Noteworthy is the incorporation of references indicating the dual usage of datasets—initially introduced for a specific task and subsequently repurposed, signified by citations in the \textit{Application} column. Regrettably, several articles utilized private datasets, although some authors may offer dataset-sharing options. We recommend consulting the corresponding articles to explore potential data access avenues.

A number of key considerations can be gleaned from Table \ref{tab:datasets}, as follows:

\begin{itemize}
    \item In the context of datasets tailored for pose estimation, a salient observation is the standardization of skeletal structures when deployed across diverse animal species~\cite{cao2019cross, yu2021ap10k, ng2022animal}. This standardization facilitates the training of a singular network, avoiding the need for species-specific networks. Conversely, datasets exclusive to individual animal species exhibit intricate skeletal configurations tailored to the anatomical nuances of that species. For instance, precision in detecting the distal and proximal ends of crickets' antennae may be achievable~\cite{fazzari2022workflow}, but such granularity may not translate to not insect species like horses.

    \item The inclination of animals like goats and birds to move in open fields compels researchers to rely on sensor data or alternative methods, such as audio recordings, for event and action detection. This approach is significantly more manageable than tracking the animals with cameras. However, the utilization of sensors is constrained by the availability and affordability of these devices, directly impacting the number of individuals involved and the quantity of sequences that can be compiled for the dataset.

    \item For identifying static positions, such as whether an animal is lying down or standing, still images suffice. However, capturing and analyzing videos, or more precisely, short video clips, is crucial for recognizing dynamic actions and behaviors. These clips are intentionally brief to focus solely on the relevant action event, ensuring accurate classification using deep learning techniques. This approach effectively eliminates extraneous or unrelated behaviors that may interfere with the identification of the specific behavioral instance. This is the rationale behind Yang \textit{et al.}'s decision to utilize 15 frames for each video clip~\cite{yang2022apt}.

    \item Unfortunately, publicly available collective and social behavior prediction and analysis datasets are currently limited to mice and fish, even though we discussed a study in this survey that utilized explainable artificial intelligence to analyze ant behavior. This innovative research methodology relies heavily on video data, presenting a computational challenge that demands substantial processing efforts. The intricate nature of this approach necessitates a considerable investment of time for meticulous frame and event labeling, thereby slightly diminishing its overall research appeal and popularity.

    \item An essential consideration pertains to the primary focus of many databases, which primarily aim at identification, detection, pose estimation, and tracking. Despite this orientation, it is crucial to acknowledge that several datasets have been instrumental in behavioral analysis, even if not explicitly designed for such purposes. Researchers are strongly encouraged not to overlook animal datasets merely because they do not pertain to specific behaviors. Valuable insights can be gleaned from these datasets, and their broader applicability should be explored beyond their initially intended scope.
\end{itemize}

\begin{table}
    \caption{Datasets Information useful for Animal Behavior Analysis}
    \resizebox{\textwidth}{!}{
    \begin{tabular}{lllll}
        \toprule
        \textbf{Species} & \textbf{Introduced by} & \textbf{Type} & \textbf{Applications} & \textbf{Dimensions} \\
        \midrule
        Birds & Ak{\c{c}}ay~\cite{akccay2020automated} & Images & Detection & 3436 images\\
        Birds & Morfi et al.~\cite{morfi2019nips4bplus} & Audio & Recognition~\cite{bravo2021bioacoustic} & 687 recording, 87 classes\\
        Birds & Knight et al.~\cite{knight2019classification} & Audio & Event detection~\cite{lostanlen2019long} & 64 recording, 5 classes\\
        Birds & Shamoun-Baranes et al.~\cite{shamoun2017short} & Sensor data & Movement prediction~\cite{wijeyakulasuriya2020machine} & 19 sequences\\
        
        Cattle & Arablouei et al.~\cite{arablouei2023multimodal} & Sensor data & Behavior classification & 11962 labeled datapoints (arm20c), 10879 labeled datapoints (arm20e)\\
        
        Dogs & Barnard et al.~\cite{barnard2016quick} & Images & Pose Estimation & 22479 images\\
        
        Goats & Kamminga et al.~\cite{kamminga2018robust} & Sensor data & Action recognition (also ~\cite{bocaj2020benefits}) & 177.8 hours of sequence data, 5 individuals\\
        Goats & Kleanthous et al.~\cite{kleanthous2022deep} & Sensor data & Action recognition (also ~\cite{bocaj2020benefits, mekruksavanich2022resnet}) & 2 sequences\\
        
        Horses & Kamminga et al.~\cite{kamminga2019dataset} & Videos & Pose estimation, Action recognition~\cite{bocaj2020benefits} & 8144 frames\\
        Horses & Mathis et al.~\cite{mathis2021pretraining} & Images & Pose Estimation &  608550 images\\
        
        Fish & Mathis et al.~\cite{mathis2018deeplabcut} & Videos & Pose estimation & 100 frames\\
        Fish & McIntosh et al.~\cite{mcintosh2020movement} & Videos & Behavior classification & 892 clips\\
        Fish & Papaspyros et al.~\cite{papaspyros2023predicting} & Videos & Collective Behavior Prediction & three 16-hour trajectory datasets\\ 
        Fish & Rahman et al.~\cite{RAHMAN2014574} & Videos & Action recognition~\cite{gore2023zebrafish} & 95 clips\\ 
        Fish & Tucker et al.~\cite{tucker2022novel} & Videos & Tracking, Chemical response analysis~\cite{gore2023zebrafish} & 3 hours for each of the 384 individuals\\ 
        
        Insects & Bjerge et al.~\cite{bjerge2023accurate} & Images & Detection & 29960 images\\
        Insects & Fazzari et al.~\cite{fazzari2} & Videos & Tracking, Chemical response analysis & 5 minutes clip for 69 individuals\\
        Insects & Modlmeier et al.~\cite{modlmeier2019ant} & Images & Movement prediction~\cite{wijeyakulasuriya2020machine} & 14400 frames (4 hours)\\
        Insects & Pereira et al.~\cite{pereira2022sleap} & Videos & Pose estimation & 30 videos, 2000 labels\\
        Insects & Pham et al.~\cite{pham2022classification} & Sequences & Locomotion classification & 258 traces\\
        Insects & Ullah et al.~\cite{ullah2022efficient} & Images & Recognition & 1686 images\\
        
        Jellyfish & Martin et al.~\cite{martin2020jellytoring} & Images & Detection & 842 images\\
        
        Mice & Burgos et al.~\cite{burgos2012social} & Videos & Pose estimation, Social behavior analysis~\cite{jiang2021multi} & 237 videos and over 8M frames\\
        Mice & Jiang et al.~\cite{jiang2021multi} & Videos & Social behavior analysis & 12*3 annotated videos, 216,000*3 frames in total\\
        Mice & Jiang et al.~\cite{jiang2022detecting} & Videos & Detection, Tracking & 10 videos, each video lasts 3 min, 4000 frames annotated\\
        Mice & Mathis et al.~\cite{mathis2018deeplabcut} & Videos & Pose estimation & 161 frames\\
        Mice & Pereira et al.~\cite{pereira2022sleap} & Videos & Pose estimation & 1000 frames, 2950 instances \& 1474 frames, 2948 instances\\
        Mice & Segalin et al.~\cite{segalin2021mouse} & Videos & Pose estimation, Behavior classification & 3.3M labels for pose annotation, 14 hours of behavior annotation\\
        
        Multiple, 5 & Cao et al.~\cite{cao2019cross} & Images & Pose Estimation & 4666 images\\
        Multiple, 30 & Yang et al.~\cite{yang2022apt} & Images/Videos & Pose Estimation & 2.4K video clips with 15 frames for each video, resulting in 36K frames\\
        Multiple, 54 & Yu et al.~\cite{yu2021ap10k} & Images & Pose Estimation & 10015 images\\
        Multiple, 850 & Ng et al.~\cite{ng2022animal} & Images/Videos & Pose Estimation, Action recognition & 50 hours of annotated videos, 30K video sequences for AAR, 33K frames for APE\\
        
        Pigs & Rielert et al.~\cite{RIEKERT2020105391} & Images & Position classification & 7277 images\\
        
        Tigers & Li et al.~\cite{li2019atrw} & Images & Pose estimation & 8076 images\\
        
        Monkeys & Labuguen et al.~\cite{labuguen2021macaquepose} & Images & Pose estimation & 13083 images\\
        \bottomrule
    \end{tabular}}
    \label{tab:datasets}
\end{table}

\section{Discussion about research questions}
In concluding our extensive survey, we undertake the task of responding to the research questions outlined in \autoref{ssec:rq}, drawing upon the insights gleaned from the studies expounded upon earlier. These responses aim to offer readers practical guidance in navigating the dynamic trends discerned from the comprehensive examination of the state-of-the-art literature. Their purpose is to serve as a compass for readers, facilitating a deeper comprehension of emerging patterns and fostering the implementation of advancements in the field of animal behavior.

\textbf{RQ1 (Which animal species are less considered and why?):}
In the context of animal behavior analysis employing deep learning, farm animals take center stage, as illustrated in \autoref{fig:analysis}. Pigs and cows emerge as prominent subjects, while mice claim a noteworthy position owing to their significance in neuroscience research~\cite{bryda2013mighty}. Despite chickens being the most widely farmed animals globally, surpassing even cows, sheep, and goats~\cite{robinson2014mapping}, their consideration in behavior analysis appears relatively subdued. This discrepancy may be attributed to the limited nature of chicken behavior, encompassing mostly activities such as sleeping, moving, and eating. They are predominantly analyzed for welfare-related assessments~\cite{mohialdin2023chicken, joo2022birds}.

Goats and sheep, less scrutinized in behavioral studies, likely owe their lower visibility to their outdoor grazing habits, which hinder the feasibility of employing image processing techniques. Unlike pigs, which are often studied within confined spaces, the vast open areas in which goats and sheep are typically raised limit the practicality of utilizing image processing, even with occasional drone deployment~\cite{al2020drones}. Birds face a similar challenge, requiring continuous tracking or data collection from sensors for comprehensive analysis~\cite{bergen2022classifying}.

For aquatic creatures like fishes, a distinct hurdle arises from the difficulty in training neural networks on underwater images. These images often suffer from poor quality due to distortion and color/contrast loss in water, necessitating an image enhancement phase for meaningful analysis~\cite{saleh2022adaptive}.

A notable observation is the limited consideration given to domestic animals in this research context~\cite{choi2021cat, lecomte2021validation, chambers2021deep, kasnesis2022deep}. This may stem from the scarcity of veterinary professionals engaged in this evolving field or ethical concerns surrounding the study of domestic animals.

\begin{figure}
    \centering
    \includegraphics[width=0.8\linewidth]{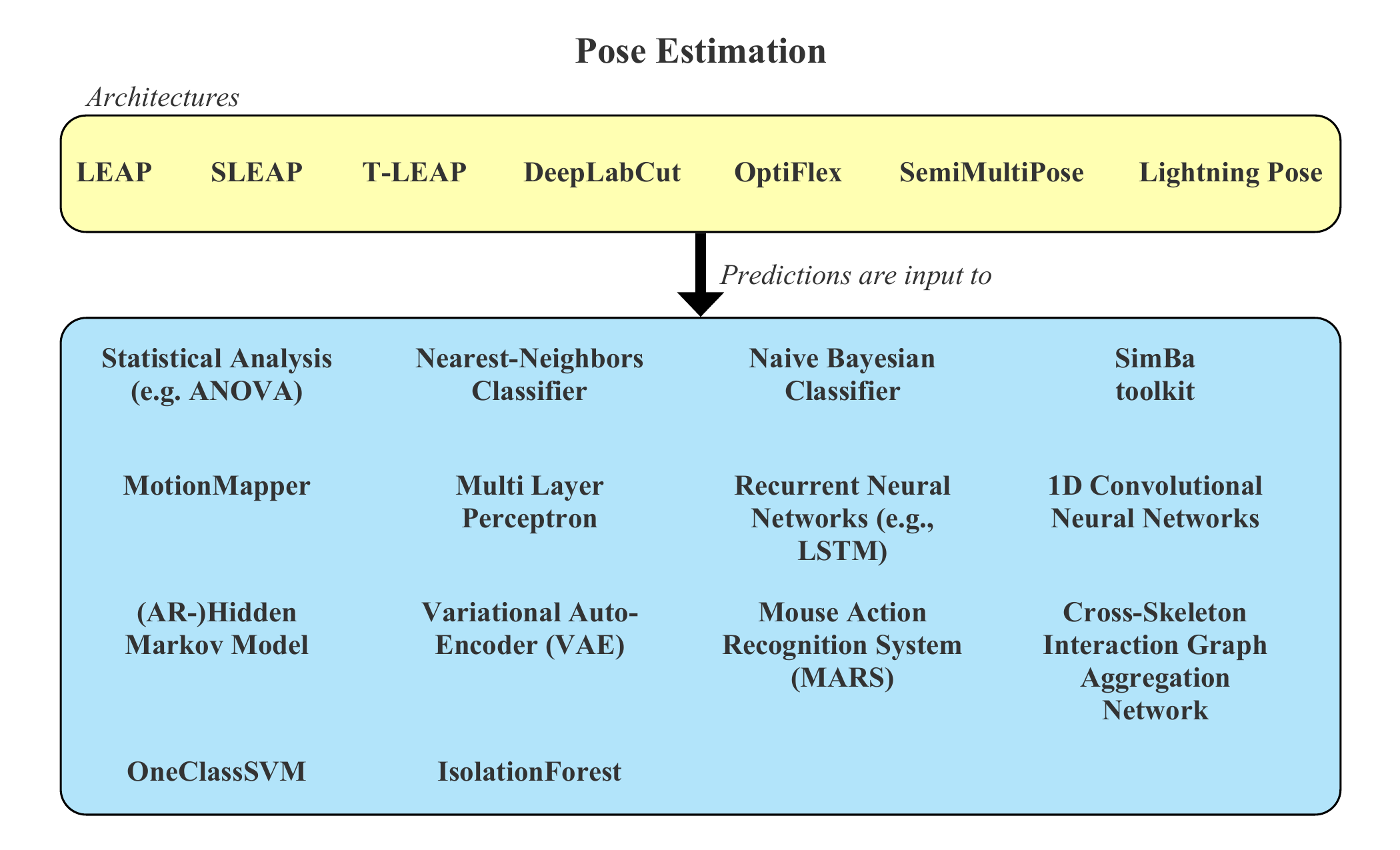}
    \caption{Comprehensive schema illustrating the pose estimation architectures covered in this survey, accompanied by detailed methodologies for accurate classification of predictions into distinct behavioral classes.}
    \label{fig:pemethods}
\end{figure}

\begin{figure}
    \centering
    \includegraphics[width=0.8\linewidth]{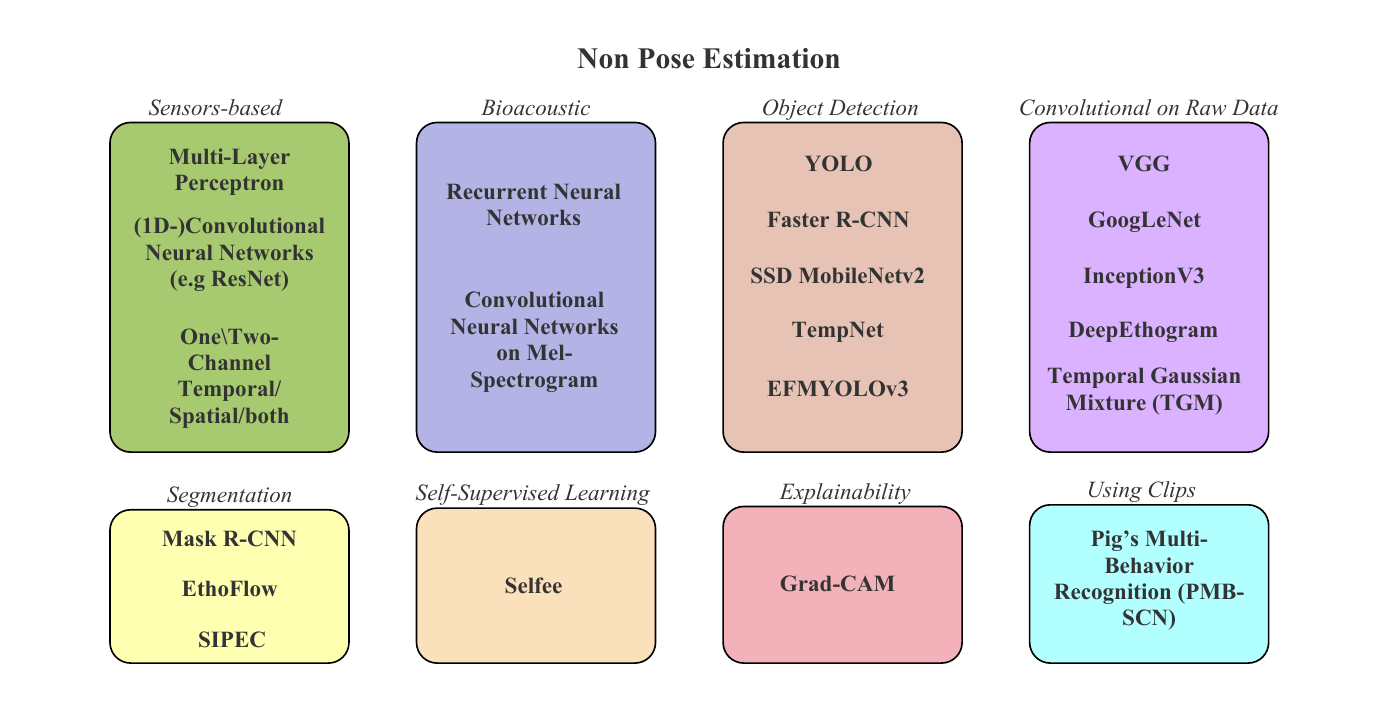}
    \caption{Comprehensive schema illustrating the non-pose estimation architectures covered in this survey. To improve readability we divided them into blocks using the same structure employed in the survey.}
    \label{fig:nonpemethods}
\end{figure}

\textbf{RQ2 (What deep learning methods have been used in the literature for animal behavior analysis?):}
Throughout this survey, we delineate two distinct methodologies employed for the analysis of animal behaviors: pose estimation and non-pose estimation methods. Pose estimation-based approaches hinge on the analysis of trajectories traced by keypoints, with subsequent utilization of various machine and deep learning techniques to scrutinize behavior. Predominantly, recurrent neural networks and 1D convolutional neural networks are the favored deep learning methods, though recent applications have also embraced variational auto-encoders for unsupervised motif identification~\cite{kingma2013auto, luxem2022identifying} and convolutional graph networks for interaction analysis~\cite{zhou2022cross}. However, a prevalent trend emerges wherein data is often processed through statistical analysis or traditional machine learning methods \cite{saleh2023pharmacology, fang2021pose, nilsson2020simple, mckenzie2023capturing}. This inclination may stem from the fact that many researchers are not inherently engaged in artificial intelligence or data science, as mentioned earlier (see \autoref{ssec:mappinganalysis}), or it could be influenced by the volume of available data, given the heightened data requirements of deep learning~\cite{ozdacs2023classification}. Despite the migration of classification tasks and behavior discovery to deep learning, certain aspects, such as behavior outlier detection, persist in employing classical machine learning approaches like OneClassSVM~\cite{boser1992training} and IsolationForest~\cite{liu2008isolation}. On the other hand, non-pose estimation encompasses diverse applications, categorized based on the type of data: sensors, audio and video, and image data. Sensor data commonly undergoes processing through (1D)-convolutional or recurrent neural networks~\cite{eerdekens2020resampling, mekruksavanich2022resnet, pan2023cnn, dang2022lorawan}, and in some cases, multi-layer perceptron classifiers~\cite{arablouei2023animal}, particularly when sensor data is transformed into features like velocity, angles, humidity, location, among others~\cite{arablouei2023multimodal}. In the domain of bioacoustics, recurrent neural networks or pre-trained convolutional neural networks on spectrogram images are frequently applied~\cite{wang2021identification}. For images and videos, processing methods vary according to the task at hand. They may be employed for object detection, where identification extends beyond the animal to encompass specific behaviors, typically achieved through frameworks such as YOLO~\cite{redmon2016you} and Faster R-CNN~\cite{ren2015faster}. Alternatively, pre-trained neural networks or segmentation techniques, with subsequent analysis of the segmentation mask, are utilized~\cite{xiao2023multi}. \autoref{fig:pemethods} and \autoref{fig:nonpemethods} encapsulate and illustrate the summarized pose and non-pose estimation methods for behavior analysis expounded in this survey.

\textbf{RQ3 (What are the differences between human and animal behavior analysis?):} 
Deep learning for animal behavior analysis draws inspiration from and translates methodologies used in human applications. While similarities exist in predicting final classification outputs, notable differences emerge in data handling, particularly with sensor data. Moreover, a distinct gap is evident in pose estimation methods. In the following list, we highlight these divergences that present opportunities for leveraging deep learning in animal behavior analysis:

\begin{itemize}
    \item \textbf{Pose estimation (regression).} In animal pose estimation, regression-based methodologies have not garnered widespread acclaim as observed in human applications. This divergence in popularity can likely be attributed to their utilization predating the ascendancy of heatmap-based techniques~\cite{zheng2023deep}. Conversely, within the human context, these regression approaches have demonstrated noteworthy success, particularly when seamlessly integrated with multi-task learning techniques~\cite{ruder2017overview}. By facilitating the exchange of information between interconnected tasks, such as pose estimation and pose-based action recognition, models can enhance their generalization capabilities. For example, Fan \textit{et al}.~\cite{fan2015combining} introduced an innovative dual-source convolutional neural network, employing both image patches and complete images for two distinct tasks: joint detection, responsible for discerning whether a patch contains a body joint, and joint localization, aimed at pinpointing the precise location of the joint within the patch. Each task is associated with its dedicated loss function, and the synergistic combination of these tasks yields notable improvements in overall performance.
    
    \item \textbf{Pose estimation (heatmap).} While these approaches find application in the realm of animal behavior, we believe that leveraging the following methods employed in human pose estimation could significantly augment this field:
        \subitem \textit{Using pre-trained an general models based on transformers.} Xu \textit{et al.}~\cite{xu2023vitpose} introduced VitPose++, an extension of the VitPose model~\cite{xu2022vitpose}, specifically designed for generic body pose estimation. This innovative framework utilizes vision transformers and distinguishes itself by not only focusing on human subjects but also extending its application to animals. The training process incorporates the AP-10K~\cite{yu2021ap} and APT36K~\cite{yang2022apt} datasets, ensuring a comprehensive understanding of both human and animal poses. We believe that fine-tuning pre-trained models of this kind may improve animal pose estimation results. 
        \subitem \textit{Using Generative Adversarial Networks (GANs)~\cite{goodfellow2014generative}.} The benefit can be twofold. GANs are explored in human pose estimation to generate biologically plausible pose configurations and to discriminate the predictions with high confidence from those with low confidence, which could infer the poses of the occluded body parts~\cite{chen2017adversarial, chou2018self}. A valuable aspect that could be used also in our task, solving possible occlusion issue in open-field research. Another usage is to perform adversarial data augmentation treating the pose estimation network as a discriminator and using augmentation networks as a generator to perform adversarial augmentations~\cite{peng2018jointly}.  
    
    \item \textbf{Smart handling of sensor data.} The significance of detecting human behaviors through sensor technologies has grown significantly in recent years~\cite{helmi2023human, park2023multicnn, islam2023multi}. This has prompted researchers to critically evaluate the relevance of various sensors employed in behavior classification through deep learning techniques. The central question revolves around the necessity of each sensor in contributing to accurate behavior identification and whether a contribution significance analysis can be effectively employed in this context. Addressing this concern, Li \textit{et al.}~\cite{li2023human} introduced a novel method aimed at optimizing sensor selection. Their approach involves assessing the self-information brought by the \textit{j}th sensor concerning the occurrence of a specific activity $A_i$ and multiplying it by the universality of the same sensor \textit{j} during instances of the same activity $A_i$. This innovative strategy proved to be highly effective, resulting in improved recognition rates and reduced time consumption, thanks to the  elimination of redundant and noisy data.
    
    \item \textbf{Quantifying data scarcity.} As highlighted in the limitations section (refer to \autoref{ssec:limit}), the collection of behavior data is frequently a laborious and resource-intensive process. Consequently, estimating dataset size has garnered attention in the field of Human Action Recognition (HAR) as an integral component of data collection planning. The objective is to mitigate the time and effort invested in behavior data collection while ensuring precise estimates of model parameters. For example, Hossain \textit{et al.}~\cite{hossain2023bayesian} introduced a method grounded in Uncertainty Quantification (UQ)~\cite{abdar2021review} to determine the optimal amount of behavior data required for obtaining accurate estimates of model parameters when modeling human behaviors as a Markov Decision Process~\cite{bellman1957}. Technologies of this nature have the potential to expedite research endeavors by facilitating more efficient workflows.

\end{itemize}

\textbf{RQ4 (What are the deep learning strategies that are suitable and could enhance this task, but are not yet exploited?):} 
Exploring animal behavior through deep learning is just one facet of the multifaceted study in ethology and neuroscience. Beyond mere identification, understanding the decision-making processes and the emergence of new behaviors in animals is of paramount importance~\cite{international2021standardized}. This survey focuses on methodologies primarily centered around detecting behavioral classes or, in the case of unsupervised learning, identifying behavioral motifs. These motifs are then grouped into behavioral classes based on similarity, often with the assistance of human experts. However, animals, much like humans, exhibit dynamic changes in their behavior over time There is a growing need for methods that can efficiently capture these trial-to-trial changes, breaking them down into a learning component and a noise component~\cite{ashwood2020inferring}. This is where reinforcement learning plays a pivotal role. Not only does it allow the perception of shifts in animal behavior and provide examples of biological learning algorithms, but it also enables the emulation of animal movements. This emulation has given rise to digital twins~\cite{liu2022digital}, providing a valuable tool for a more comprehensive study of animal behavior thanks to the ease of data acquisition.
The convergence of deep learning strategies and reinforcement learning opens up exciting possibilities, paving the way for the development of interactive simulacra of animal behavior, akin to advancements already achieved in human behavior studies~\cite{park2023generative}. This innovative approach empowers researchers to simulate how animals might adapt their behavior in diverse environments while interacting with various agents, be they other animals, humans, or even robots~\cite{yamaguchi2018identification, mori2022probabilistic, romano2021animal}. Ultimately, the fusion of deep learning and reinforcement learning holds the promise of creating dynamic, interactive simulations that significantly deepen our understanding of the nuances of animal behavior across a spectrum of contexts.

\section{Conclusions}
In summary, this survey rigorously examined the manifold benefits associated with the application of deep learning methodologies in the identification of animal behavior. Commencing with a detailed elucidation of the underlying motivations, limitations, objectives, and pertinent research inquiries steering the integration of deep learning within this domain, we established a robust contextual framework. Subsequently, our scrutiny extended to a thorough review of contemporary techniques, systematically categorized into pose and non-pose estimation methodologies.
Within these delineations, we expounded upon a spectrum of methodologies, encompassing sequence processing in pose estimation and biocoustics, direct classification utilizing convolutional neural networks, outliers detection, convolutional graph neural networks, object detection, segmentation strategies, self-supervised learning, explainability, and unsupervised learning. Moreover, we curated a comprehensive table of publicly available datasets relevant to animal behavior, thereby augmenting the practical utility of deep learning applications.
Our discourse on the subject and prospective considerations has pinpointed extant challenges within the literature, proffering a roadmap for potential research trajectories conducive to the advancement of the field. In essence, this survey serves as an invaluable compendium for researchers spanning diverse domains, with particular relevance to ethologists and neuroscientists. We believe that this survey will function as a guiding beacon, steering forthcoming research initiatives and fostering advancements in the intricate domain of animal behavior studies using deep learning.

\bibliographystyle{ACM-Reference-Format}
\bibliography{sample-base}

\end{document}